\documentclass[times,referee,twocolumn,final,authoryear]{elsarticle}

\usepackage{ycviu}
\usepackage{framed,multirow}
\usepackage{multicol}
\usepackage{svg}
\usepackage{latexsym}

\usepackage{xcolor}
\usepackage{graphicx}
\usepackage{amsmath,amssymb} 
\usepackage{comment}
\usepackage{booktabs}
\usepackage{tabularx}
\usepackage{xspace}
\usepackage{soul}
\usepackage{url}
\usepackage{enumitem}
\usepackage{caption}
\usepackage{multirow}
\usepackage{algorithm}
\usepackage{algpseudocode}
\usepackage{subcaption}
\usepackage{makecell}

\newcommand{\Linear}{\text{Linear}}

\newcommand{\bmx}[0]{\begin{bmatrix}}
\newcommand{\emx}[0]{\end{bmatrix}}

\newcommand{\E}[0]{\mathbb{E}}

\newcolumntype{?}{!{\vrule width 0.8pt}}

\journal{Computer Vision and Image Understanding}

\begin{document}

\begin{frontmatter}

\title{AC-VRNN: Attentive Conditional-VRNN for Multi-Future Trajectory Prediction}

\author[1]{Alessia \snm{Bertugli}\corref{cor1}}
\cortext[cor1]{Corresponding author:}
\ead{alessia.bertugli@unitn.it}
\author[2]{Simone \snm{Calderara}}
\author[3]{Pasquale \snm{Coscia}}
\author[3]{Lamberto \snm{Ballan}}
\author[2]{Rita \snm{Cucchiara}}

\address[1]{Dipartimento di Ingegneria e Scienza dell'Informazione, Universit\'a degli Studi di Trento, Italy}
\address[2]{Dipartimento di Ingegneria ``Enzo Ferrari'', Universit\'a degli Studi di Modena e Reggio Emilia,  Italy}
\address[3]{Dipartimento di Matematica “Tullio Levi-Civita”, Universit\'a degli Studi di Padova,  Italy}

\received{1 May 2013}
\finalform{10 May 2013}
\accepted{13 May 2013}
\availableonline{15 May 2013}
\communicated{S. Sarkar}

\begin{abstract}
Anticipating human motion in crowded scenarios is essential for developing intelligent transportation systems, social-aware robots and advanced video surveillance applications. A key component of this task is represented by the inherently multi-modal nature of human paths which makes socially acceptable multiple futures when human interactions are involved.
To this end, we propose a generative architecture for multi-future trajectory predictions based on Conditional Variational Recurrent Neural Networks (C-VRNNs). Conditioning mainly relies on prior belief maps, representing most likely moving directions and forcing the model to consider past observed dynamics in generating future positions.
Human interactions are modeled with a graph-based attention mechanism enabling an online attentive hidden state refinement of the recurrent estimation. 
To corroborate our model, we perform extensive experiments on publicly-available datasets (e.g., ETH/UCY, Stanford Drone Dataset, STATS SportVU NBA, Intersection Drone Dataset and TrajNet++) and demonstrate its effectiveness in crowded scenes compared to several state-of-the-art methods. \\ \\
Pre-print, doi: \url{https://doi.org/10.1016/j.cviu.2021.103245} \\ 
Received 24 July 2020; Received in revised form 24 June 2021; Accepted 30 June 2021 \\
Available online xxxx1077-3142/©2021 Published by Elsevier Inc. \\
\end{abstract}



\end{frontmatter}

\section{Introduction}
Trajectory forecasting has recently experienced exponential growth in several research areas such as video surveillance, sports analytics, self-driving cars and physical systems (\cite{trajsurvey}). 
Its main applications include pedestrians dynamics prediction (\cite{socialforce,slstm,sslstm,sgan,stgat,srlstm}), vehicles behaviour analysis (\cite{Jiachen_IROS19,trafficpredict,multi-pred,desire}) as well as intent estimation of people and cars on roads to avoid possible crashes. In sports analytics (\cite{stats,graphvrnn,weeksup,stocpred,Chen2018GeneratingDP,Hsieh_2019}), being able to predict players trajectories can improve the action interpretation of each player while in physical systems it can be fundamental to predict particles dynamics in complex domains (\cite{nri,meta-nri,fact-nri}).

In this paper, we focus on predicting human dynamics in crowded contexts (e.g., shop entrances, university campuses and intersections)  where people and autonomous vehicles mainly manifest their complex and multi-modal nature. Typically, two different strategies are employed to model human interactions: \emph{pooling-based} and \emph{graph-based} methods. Pooling-based methods (\cite{slstm,sslstm,sgan,socialways,peekingintothefuture,socialas}) employ sequence-to-sequence models to extract features and generate subsequent time steps, interspersed with pooling layers to model interactions between neighbours.
By contrast, graph-based methods (\cite{stgat,srlstm,trafficpredict,graphvrnn,stocpred,socialattention,garden}) apply graph neural networks to model interactions. Although these approaches have proven to be effective, some problems are still open, such as efficiently exploiting context cues and appropriately capturing human interactions in critical situations. Another relevant aspect to consider in trajectory prediction is represented by scene constraints like walls and other obstacles which strongly influence human motion. A common approach to overcome this issue is to introduce visual elements into the network such as images or semantic segmentation (\cite{peekingintothefuture,social-bigat,sophie}) yet this implies the availability of video streams both at train and test time.
To this end, we propose a novel method for multi-future trajectory forecasting that works in a completely generative setting, enabling the prediction of multiple possible futures. 
During online inference, we integrate human interactions at time step level, allowing other agents to affect the whole trajectory generation process. As a consequence, online interactions computation improves the predicted trajectories as the number of time steps increases 
limiting the error growth. To take into account past human motion, local belief maps steer future positions towards more frequent crossed areas when human interactions are limited or absent. Technically, our model is a Conditional-VRNN, conditioned by prior belief maps on pedestrians frequent paths, that predicts future positions one time step at a time, by relying on recurrent network hidden states refined with an attention-based mechanism.

The main contributions of this paper are two-fold:
\begin{enumerate}[label=(\roman*)]
    \item We propose a novel method to integrate human interactions into the model in an online fashion, relying on a hidden state refinement process with a graph attentive mechanism. We employ a similarity-based adjacency matrix to take into account pedestrians' neighbourhoods.
    \item We introduce local belief maps to encourage the model to follow a prior transition distribution whenever the prediction is uncertain and to discourage unnatural behaviour such as crossing obstacles, avoiding employing additional visual inputs. In this way, future positions may take advantage of prior knowledge while being predicted. Such behaviour is imposed during training by a Kullback–Leibler (KL) divergence loss between ground-truths and samples contributing to the model performance refinement.
\end{enumerate}
We demonstrate that our model achieves state-of-the-art performance on several standard benchmarks using different evaluation protocols. We also outperform our competitors on the challenging Stanford Drone Dataset (SDD) and the recent Intersection Drone Dataset (InD) showing the robustness of our architecture to more complex urban contexts. Furthermore, we test our model on human dynamics collected from basketball players to analyze its ability to capture complex interactions in confined areas. Finally, our architecture positions among the best models on the TrajNet++ benchmark.
\section{Related Work}
Traditionally, trajectory prediction has been approached with rule-based and social force models (\cite{socialforce,socialforce2,socialforce3}) that have been proven to be effective in simple contexts, but fail to generalize to complex domains.
In recent years, generative models (\cite{sgan,weeksup,stocpred,social-bigat,trajectron}) have been focusing on the multi-modal nature of this task since multiple human paths could be regarded as socially acceptable despite being different from ground-truth annotations.
In the following, we group related work into position-based models, which uses only spatial information, and graph-based models, which rely on connected structures.

\textbf{Position-based models.} Social-LSTM~(\cite{slstm}) models individual trajectory as a long short-term memory (LSTM) encoder-decoder and considers interactions using a social pooling mechanism. 
Social GAN~(\cite{sgan}) uses a pooling mechanism in combination with a generative model to predict socially acceptable trajectories. SoPhie~(\cite{sophie}) consists of a Generative Adversarial Network (GAN), which leverages the contribution of a social attention module and a physical attention module. 
SS-LSTM~(\cite{sslstm}) uses different inputs to also take into account the influence of the environment and maps of the neighbourhood to narrow the field of mutual influences.

\textbf{Graph-based models.} Graph Neural Networks (GNNs) have been used to model interactions between different trajectories. 
Graph Variational RNNs~(\cite{graphvrnn}) model multi-agent trajectory data mainly focusing on multi-player sports games.
Each agent is represented by a VRNN where prior, encoder and decoder are modeled as message passing GNNs, allowing the agents to weakly share information through nodes.
Graph-structured VRNN 
network~(\cite{stocpred}), based on relation networks, infers the current state and forecasts future states of basket and football players trajectories. SR-LSTM~(\cite{srlstm}) uses a state refinement module through a motion gate and pedestrian-wise attention. Social-BiGAT~(\cite{social-bigat}) presents a graph-based generative adversarial network based on GAT~(\cite{gat}) that learns reliable future representations that encode the social interactions between humans in the scene and contextual images to incorporate scene information. 
ST-GAT~(\cite{stgat}) proposes a model based on two levels of LSTMs to incorporate interactions through a hidden state refinement, which uses a GAT in the encoding part, while the decoder generates future positions.

Compared to approaches based on Variational Autoencoders (VAEs)~(\cite{graphvrnn,stocpred,nri}), our method does not model the prediction as a graph yet uses an attentive module to refine the hidden state of a recurrent network. Doing so, information about other agents influences the prediction. Unlike sports games, where all players share the same goal, in urban scenarios neighbourhood information is crucial since future paths may depend on mutual distances among people.
Our model resembles SR-LSTM~(\cite{srlstm}) and STGAT~(\cite{stgat}) combining LSTMs and GNNs. Nevertheless, SR-LSTM~(\cite{srlstm}) exploits cell states of LSTMs limiting the observation horizon. STGAT~(\cite{stgat}) uses GAT~(\cite{gat}) as hidden state refinement, but it employes a sequence-to-sequence model without an online refinement. Both methods do not take into account contextual information or collective behaviors (e.g., belief maps) in order to avoid uncommon paths.
\section{AC-VRNN Model}
Pedestrian dynamics are primarily affected by the neighbourhood space in urban areas. To avoid obstacles or other people, pedestrians continuously steer their motion gaining also the advantage of prior knowledge acquired in similar contexts. To this end, our model relies on past motions of monitored scenes as well as structured interactions in a generative setting.

\textbf{Problem formulation.} Given a pedestrian at time step $t$, his/her current position is represented by 2-D coordinates. Our model analyzes 
$T_{obs}$ time steps to predict motion dynamics during the next $T_{pred}$ time steps. Similarly to \citet{sgan}, our model uses displacements with respect to the previous points. More specifically, given a sequence of displacements ($ x_{0}, .., x_{T_{pred}}$), we observe a part of the sequence ($x_{0}, ..., x_{T_{obs}}$) and predict the subsequent one ($x_{T_{obs}+1}, ..., x_{T_{pred}}$). 

\begin{figure*}[!t]
\centering
\includegraphics[height=5.1cm]{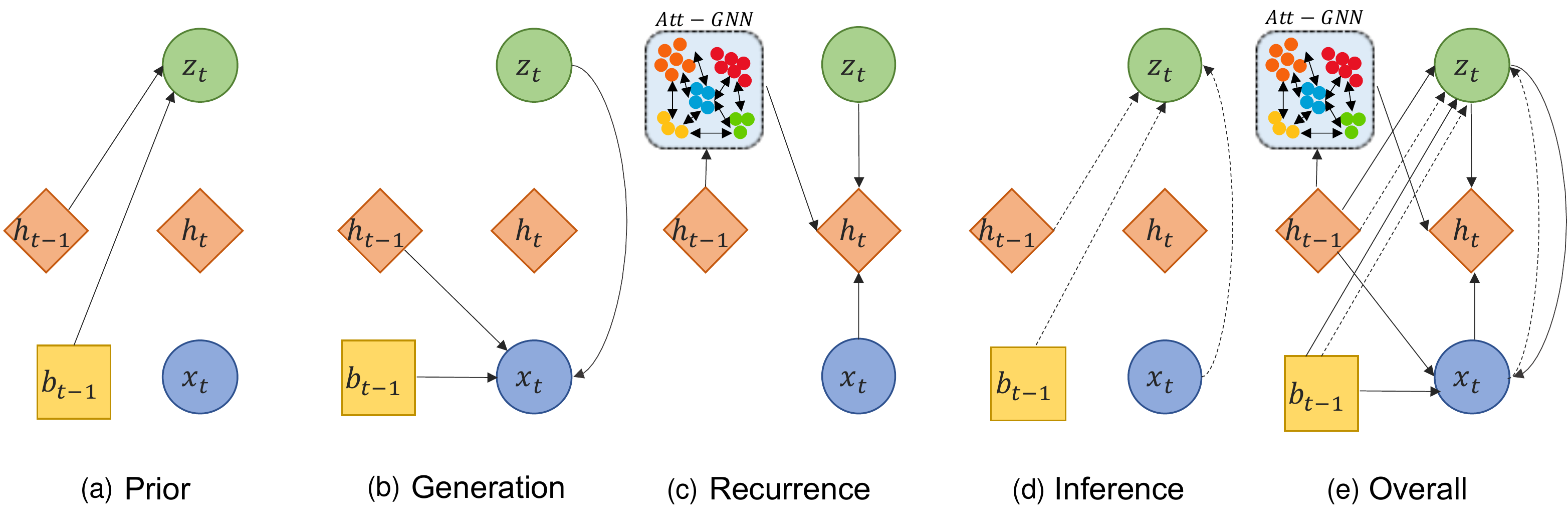}
\caption{Illustration of each phase of our AC-VRNN architecture for a time step $t$. 
A recurrent variational autoencoder is conditioned on prior belief maps $b_{t-1}$. The hidden state of the RNN $h_{t-1}$ is refined with an attentive module obtaining $h'_t$, that replaces $h_t$ in the next step of recurrence. At inference time, it generates future displacements using the prior network on $\mathbf{h}_{t}$ and makes an online computation of the adjacency matrix which defines connections between pairs of nodes.}
\label{fig:model}
\end{figure*}

Our Attentive Conditional Variational Recurrent Neural Network (AC-VRNN) is composed of three building blocks: (i) a VRNN to generate a sequences of displacements in a multi-modal way; (ii) a hidden state refinement based on an attentive mechanism to model the interactions within the neighbourhood, performed at a time step level during training and inference phases; (iii) a belief map to encourage the model to follow prior belief maps when it is uncertain, avoiding predicting samples that may fall within never crossed areas. A complete illustration of all phases of AC-VRNN is shown in Figure~\ref{fig:model}.


\textbf{Predictive VRNN.} 
VRNNs~(\cite{vrnn}) explicitly model dependencies between latent random variables $\mathbf{z}_{t}$ across subsequent time steps. They contain a Variational Autoencoder (VAE)~(\cite{vae}) at each time step conditioned on the hidden state variable $\mathbf{h}_{t-1}$ of an RNN to take into account temporal structures of sequential data. At each time step, prior, encoder and decoder output multivariate normal distributions, with three functions ($f_{\mathrm{pri}}$, $f_{\mathrm{enc}}$ and $f_{\mathrm{dec}}$) modeling their means and variances.
Since the true posterior is intractable, it is approximated by a neural network $q_{\phi}$, which also depends on the hidden state $\mathbf{h}_{t-1}$ under recurrency equations as follows:
\begin{align}
    \label{eq:prior}
    p_{\theta}\left(\mathbf{z}_{t} | \mathbf{x}_{<t}, \mathbf{z}_{<t}\right) &=\mathcal{N}\left(\mathbf{z}_{t} | \boldsymbol{\mu}_{\mathrm{pri,t}},\left(\boldsymbol{\sigma}_{\text {pri,t}}\right)^{2}\right), & \text{(prior)} \\
    \label{eq:inference}
    q_{\phi}\left(\mathbf{z}_{t} | \mathbf{x}_{\leq t}, \mathbf{z}_{<t}\right) &=\mathcal{N}\left(\mathbf{z}_{t} | \boldsymbol{\mu}_{\mathrm{enc,t}},\left(\boldsymbol{\sigma}_{\text {enc,t }}\right)^{2}\right), & \text{(inference)} \\
    \label{eq:generation}
    p_{\theta}\left(\mathbf{x}_{t} | \mathbf{x}_{<t}, \mathbf{z}_{\leq t}\right) &=\mathcal{N}\left(\mathbf{x}_{t} | \boldsymbol{\mu}_{\mathrm{dec,t}},\left(\boldsymbol{\sigma}_{\text{dec,t}}\right)^{2}\right), & \text{(generation)} \\
    \label{eq:recurrence}
    \mathbf{h}_{t} &=f_{\mathrm{rnn}}\left(\mathbf{x}_{t}, \mathbf{z}_{t}, \mathbf{h}_{t-1}\right). & \text{(recurrence)}
\end{align}
These functions can be deep neural networks with learnable parameters $\theta$ and $\phi$ that output $\left(\boldsymbol{\mu}_{\mathrm{pri,t}}, \boldsymbol{\sigma}_{\mathrm{pri,t}} \right)$,
$\left(\boldsymbol{\mu}_{\mathrm{enc,t}}, \boldsymbol{\sigma}_{\mathrm{enc,t}} \right)$ and $\left(\boldsymbol{\mu}_{\mathrm{dec,t}},\boldsymbol{\sigma}_{\mathrm{dec,t}} \right)$, respectively.
The generative and inference processes are jointly optimized by maximizing the following variational lower bound (ELBO) with respect to their parameters\footnote{In order to keep the notation light we omit the conditioning variables.}:

\begin{align}
    \label{eq:objective_function}
     ELBO = \E_{q_{\phi,t}(\mathbf{z}_t)} \left[\sum_{t=1}^{T}\left(-\mathrm{KL}\left(q_{\phi, t}(\mathbf{z}_t) \| p_{\theta, t}(\mathbf{z}_t) \right) + \log p_{\theta, t}(\mathbf{x}_t) \right) \right].
\end{align}
VRNNs are typically employed to generate sequences from scratch, in a fully generative setting. However, our task is to imitate training data rather than generate completely new data at evaluation time. In a predictive setting, the predicted positions must rely on the observed ones; without any information coming from the past, future positions would only be random. Using a fully generative setting, the model would not have any chance to exploit previous observations.
For this reason, we have modified the inference protocol to generate sequences using the hidden state of the last observed time step.
VRNN learns at each time step to generate the current displacement, given the input and the RNN's hidden state. At inference time, the model only uses the last hidden state from the observed sequence, then generates the subsequent time step.
For the above reasons, AC-VRNN is a generative model used in a predictive setting: it generates one displacement at a time and becomes easy to embed human interactions at time step level.

\textbf{Attentive Hidden State Refinement.}
Pedestrians dynamics are mainly influenced by surrounding agents. Our model handles human interactions using an attentive hidden state refinement of our recurrent network through a graph neural network, as shown in Figure~\ref{fig:graph&map}(a).
Our hidden state refinement resembles the idea proposed by \cite{gat} which adopts an attention mechanism to learn relative weights between two connected nodes, through specific transformations called graph attentional layers.
At time step $t$, our refinement strategy considers a set of hidden state nodes $\{\mathbf{h}_t^1,\dots,\mathbf{h}_t^N\}$, where each $\mathbf{h}_t^i \in \mathbb{R}^F$ represents the hidden state of the $i^{th}$ agent in the scene.
The attention layer produces a new set of node features $\{\mathbf{\hat{h}}_t^1,\dots,\mathbf{\hat{h}}_t^N\}$, $\mathbf{\hat{h}}_t^i \in \mathbb{R}^{F'}$ as its output.
The transformation is parametrized by a weight matrix $\mathbf{W} \in \mathbb{R}^{F'\times F}$ (shared between graph nodes) and a weight vector $\mathbf{a} \in \mathbb{R}^{2F'}$. Self-attention coefficients $\alpha_{i, j}$ between the nodes $\mathbf{h}_t^i$ and $\mathbf{h}_t^j$ are computed as follows:
\begin{align}
\label{eq:coeff}
\alpha_{i, j}=\frac{\exp \left(\text{LeakyReLU} \left(\mathbf{a}^{T}\left[\mathbf{W} \mathbf{h}_t^i \| \mathbf{W} \mathbf{h}_t^j\right]\right)\right)}{\sum_{k \in \mathcal{N}_i} \exp \left(\text{LeakyReLU} \left(\mathbf{a}^T\left[\mathbf{W} \mathbf{h}_t^i \| \mathbf{W} \mathbf{h}_t^k\right]\right)\right)},
\end{align}
where $\|$ represents the concatenation operator.
The normalized attention coefficients are used to compute a linear combination of the features which represents the final output feature for every node, followed by a ELU non-linearity~(\cite{clevert2015fast}) acting on the neighbourhood $\mathcal{N}_i$ of the $i^{th}$ node:
\begin{align}
    \label{eq:h_att}
    \mathbf{\hat{h}}_t^i=\text{ELU}\left(\sum_{j \in \mathcal{N}_{i}} \alpha_{i, j} \mathbf{W} \mathbf{h}_t^j\right).
\end{align}
The neighbourhood $\mathcal{N}_{i}$ defines the set of nodes with positive adjacency with respect to the $i^{th}$ agent.
The adjacency matrix follows a similarity-based principle, and it is computed, inspired by proxemics interaction theory~(\cite{proxemics}), considering the heat kernel of the distance $d(i,j)$ between each pedestrian, $\exp \left( {-\frac{d(i,j)}{2\sigma^{2}}} \right),$
where $\sigma$ is a smoothing hyperparameter.
During training, our VRNN takes as input a set of sequences for a time step $t$. Then, it samples the next position $\mathbf{x}_t^i$ for each pedestrian $i$. 
Finally, the graph attention mechanism acts on the hidden state $\mathbf{h}_{t}^i$ (provided by Eq. \eqref{eq:recurrence}) to compute the corresponding interaction-refined state $\mathbf{\hat{h}}_{t}^i$.
The refined hidden state $\mathbf{\hat{h}}_{t}^i$ is concatenated to the original one and a final linear projection is applied as follows:
\begin{align}
    \label{eq:hg}
    \mathbf{h}_{t}^{'i} = \Linear \left(\mathbf{h}_{t}^i \mathbin\Vert \mathbf{\hat{h}}_{t}^i\right).
\end{align}
At the next time step, our VRNN uses the refined hidden state $\mathbf{h}_{t}^{'i}$ which carries information about interactions of previous time steps.

\textbf{Conditional-VRNN on Belief Maps.}
Since AC-VRNN is a stochastic model, it could potentially exhibit high predictive variance hence generating predictions far from expected ones. 
To balance the bias/variance trade-off of the predictor, we introduce belief maps on displacements. 
Belief maps collect data about crossed areas at training time; therefore, they contain information about the collective behavior of monitored agents. Conditioning the prediction to such maps, may lead the model to follow past behaviors and, at the same time, discourage it to predict displacements far from past crossed areas, avoiding the generation of non-realistic paths.

\begin{algorithm}[!t]
  \begin{algorithmic}[1]
    \scriptsize
    \Function{belief\_maps\_generation}{$trajectories$}
    \State  $N, M, \delta_x, \delta_y \gets get\_grid\_coarse(trajectories)$
    \State $x_{min}, y_{min}, x_{max}, y_{max} \gets get\_min\_max(trajectories)$
    \State $global\_grid \gets make\_global\_grid(x_{min}, y_{min}, N, M,\delta_x, \delta_y) $
    \ForAll{$bin \in global\_grid$}
        \State $maps \gets [0,..,0]$
        \ForAll{$trajectory \in trajectories$}
            \State $neighbour\_centres \gets get\_neighbour\_centres(bin, \delta_x, \delta_y)$
            \ForAll{$index, coord \in trajectory$}
                \If {$coord_x \in [bin_x, bin_x + \delta_x]$ \textbf{and}  $coord_y \in [bin_y, bin_y + \delta_y]$}
                \State $next\_coord \gets trajectory[index + 1]$
                \State $map \gets similarity\_matrix(next\_coord, neighbour\_centres, map)$
                \EndIf
            \EndFor
        \EndFor 
        \State $map \gets normalize(map)$ 
        \State $maps \gets insert(map)$
    \EndFor    
    \State \textbf{return} $maps$
    \EndFunction
    \Statex
    \Function{get\_grid\_coarse}{$trajectories$}
    \State $\mu_x, \mu_y \gets mean\_displacements(trajectories)$
    \State $\sigma_x, \sigma_y \gets standard\_deviation\_displacements(trajectories)$
    \State $x_{min}, y_{min}, x_{max}, y_{max} \gets get\_min\_max(trajectories)$
    \State $N \gets \frac{x_{max} - x_{min}}{\frac{\mu_x + \sigma_x}{2}};  M \gets \frac{y_{max} - y_{min}}{\frac{\mu_y + \sigma_y}{2}}$
    \State $ \delta_x \gets \frac{x_{max} - x_{min}}{N} ; \delta_y \gets \frac{y_{max} - y_{min}}{M}$
    \State \textbf{return} $N, M, \delta_x, \delta_y$
    \EndFunction
    \Statex
    \Function{similarity\_matrix}{$next\_coord, neighbour\_centres, map$}
    \ForAll{$index, centre \in neighbour\_centres$}
        \State $map[index] \gets accumulate(e^{- \sqrt{(next\_coord_{x} - centre_x)^2 + (next\_coord_{y} - centre_y)^2}})$
    \EndFor
    \State \textbf{return} $map$
    \EndFunction
  \end{algorithmic}
  \caption{Belief Maps Generation Algorithm}
  \label{alg:beliefMaps}
\end{algorithm}
Belief maps are computed dividing the coordinate space for each scene into a $N\times M$ grid. The boundaries of this grid are defined by minimum and maximum coordinates along $x$ and $y$ directions. Both past and future information of training trajectories are considered. These values could also be obtained manually defining the allowed area for predicting new coordinates. 
The values of $N$ and $M$ define the grid coarse and are computed considering the average displacement $\mu$ and its standard deviation $\sigma$ as follows:
\begin{align}
    \label{eq:bin_dim}
    N = \biggl \lfloor \frac{(x_{max} - x_{min})}{\frac{\mu + \sigma}{2}} \biggr \rfloor, \quad
    M = \biggl \lfloor \frac{(y_{max} - y_{min})}{\frac{\mu + \sigma}{2}} \biggr \rfloor.
\end{align}
For each grid location (bin), a $L\times L$ neighbourhood is then considered (with $L=5$). For each $(x,y)$ location, we get the corresponding $L\times L$ neighbourhood and compute heat kernels between the next location and the neighbourhood bins centres\footnote{$5\times 5$ belief maps along with the proposed global grid's partition guarantee that future displacements fall into the corresponding belief maps.
}. This procedure is repeated for all the trajectories and bins values are accumulated by summation.
Each belief map $\mathbf{b}$, i.e. a $L\times L$ sub-grid indexed by the $(x,y)$ location in the scene, is subsequently normalized in order to transform the cumulative grid into a probability distribution. 
Unlike the recurrent process within the VRNN, the creation of belief maps is a Markov process, as their generation only depends on single-step transitions. Details in Algorithm~\ref{alg:beliefMaps} each step for generating our belief maps.

\textbf{Conditional-VRNN.}
We exploit the belief maps to encourage the model to follow the average behaviour shown by previously observed agents. 
In our work, we use a recurrent version of CVAE~(\cite{cvae}), conditioning VRNN on belief maps. At each time step, \emph{prior}, \emph{encoder} and \emph{decoder} networks take the belief map at $t-1$ as input, conditioning the resultant Gaussian distribution. We embed belief maps with a linear projection before feeding them into the VRNN blocks:

\begin{figure*}[!t]
\centering
\begin{tabular}{c c}
\includegraphics[width=10cm]{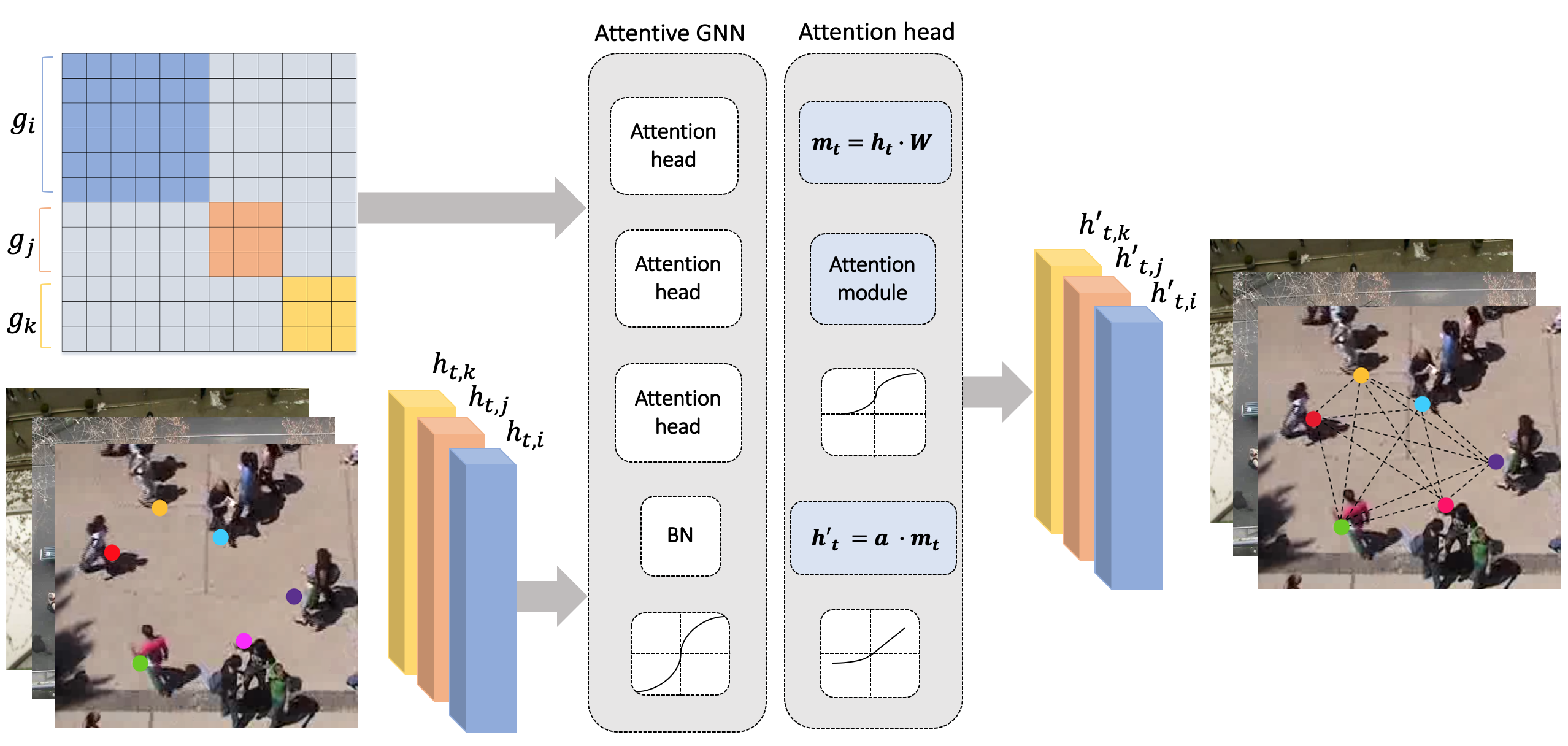} & 
\quad \includegraphics[width=7cm]{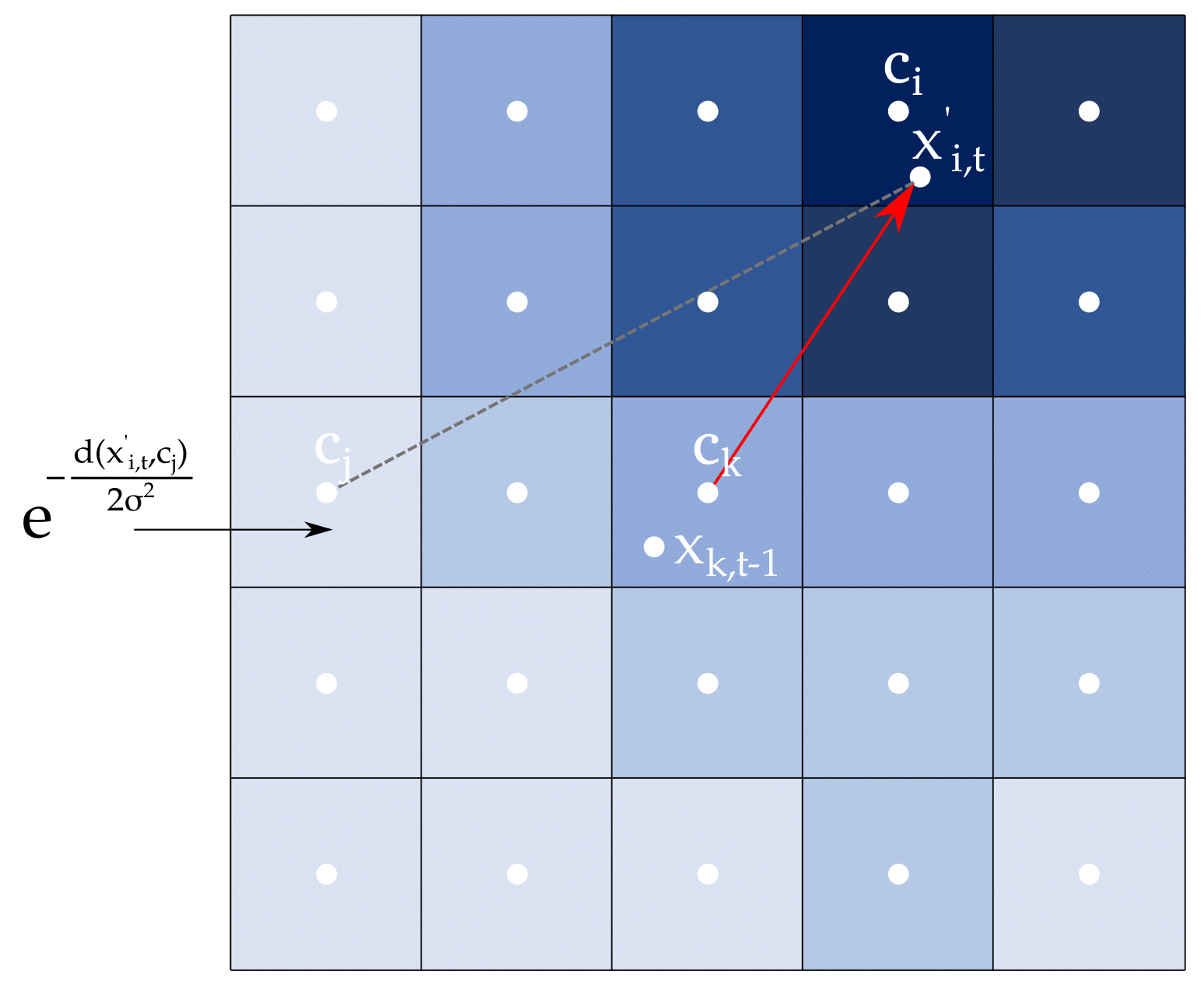}\\
(a) & (b)
\end{tabular}
\caption{Scheme of the proposed attentive hidden state refinement process. (a) The adjacency matrix is an irregular block matrix where each block size is defined by the number of pedestrians in the current scene. (b) Belief map during training for one sample using heat similarity-based strategy. The map is centred at $t-1$ to display the sampled displacements distribution at $t$.}
\label{fig:graph&map}
\end{figure*}

\begin{align}
    \label{eq:norm_pri}
    {\boldsymbol{\mu}_{\mathrm{pri,t}}, \boldsymbol{\sigma}_{\mathrm{pri,t}}=f_{\mathrm{pri}}\left(\mathbf{h}_{t-1},  \mathbf{b}_{t-1} ; \theta\right)} \\
    \label{eq:norm_enc}
    \boldsymbol{\mu}_{\mathrm{enc,t}}, {\boldsymbol{\sigma}_{\mathrm{enc,t}}=f_{\mathrm{enc}}\left(\mathbf{x}_{t}, \mathbf{h}_{t-1}, \mathbf{b}_{t-1} ; \boldsymbol{\phi}\right)} \\
    \label{eq:norm_dec}
    \boldsymbol{\mu}_{\mathrm{dec,t}}, {\boldsymbol{\sigma}_{\mathrm{dec,t}}=f_{\mathrm{dec}}\left(\mathbf{z}_{t},  \mathbf{h}_{t-1},  \mathbf{b}_{t-1} ; \theta\right)}
\end{align}
In addition to conditioning the model on belief maps, a further loss term is inserted, in order to optimize the affinity between ground-truth maps and those generated by the model. By sampling multiple displacements from the model, we obtain the sampled candidate belief map $\mathbf{b'}_{t-1}$, which identifies a probability distribution over local bin transitions. For each sampled displacement and subsequent location, we firstly index the corresponding grid bin, then the heat kernel value between the sampled next location and the $L\times L$ neighbourhood bin centres is used to fill the grid (see Figure~\ref{fig:graph&map}(b)). To build the non-ground truth belief maps, we only use the information about the position at $x_{t-1}$, and then draw $N$ samples from our model.
The aforementioned procedure allows the model to unroll the sub-grids, obtaining for every location a discrete probability density of possible transitions.
Thus, it is possible to compare generated belief maps $\mathbf{b'}_{t-1}$ and ground-truth ones $\mathbf{b}_{t-1}$ by means of the KL divergence, exploiting the histogram loss term proposed by \cite{histogram_loss}.
We add this contribution to the ELBO loss in Eq.~\eqref{eq:objective_function} encouraging the model to be compliant to the collective behaviour of all agents.
Such a divergence measure is multiplied by a constant $k$ for loss balancing to ensure that its weight is comparable to the other loss components:
\begin{equation}
\label{eq:loss}
\begin{split}
\mathcal{L} = \E_{q_{\phi, t}(\mathbf{z}_t)}\Bigl[\sum_{t=1}^{T}\Bigl(-\mathrm{KL}\left(q_{\phi, t}(\mathbf{z}_t) \| p_{\theta, t}(\mathbf{z}_t) \right) + \log p_{\theta, t}(\mathbf{x}_t) \\+ k \, \mathrm{KL}(\mathbf{b}_{t-1} \| \mathbf{b'}_{t-1})\Bigr)\Bigr].
\end{split}
\end{equation}

\section{Experiments}

\subsection{Datasets} We present experiments on different datasets to prove the robustness of our model on various scenarios and protocols. More specifically, we define multiple experiments on ETH~(\cite{pellegrini}), UCY(~\cite{ucy}), Stanford Drone Dataset~(\cite{sdd}), STATS SportVU NBA~\footnote{SportVU - STATS Perform, \url{https://www.statsperform.com/team-performance/basketball/optical-tracking/}}, Intersection Drone Dataset (inD)~(\cite{ind}), and TrajNet++~(\cite{trajnetpp}).

\textbf{ETH-UCY.}
ETH~(\cite{pellegrini}) consists of two scenes, \emph{Eth} and \emph{Hotel}, while UCY~(\cite{ucy}) consists of three scenes, \emph{Zara1}, \emph{Zara2} and \emph{Univ}. 
The benchmark contains different types of interactions among pedestrians and fixed obstacles such as buildings or parked cars.

\textbf{Stanford Drone Dataset (SDD)}~(\cite{sdd}).
SDD is a large scale dataset, containing urban scenes of a university campus, streets and intersections, shot by a drone. 
More specifically, it is composed of 31 videos of 8 different scenarios.
This dataset provides more complex scenes compared to the previous ones, involving various types of human interactions.
We use the version proposed by TrajNet benchmark~(\cite{trajnet2018,red2018}) which contains only pedestrian annotations.
We split the training set into three sets for the learning process selecting $70\%$ of data as training, $10\%$ as validation and the remaining part as testing. 

\textbf{STATS SportVU NBA}~\footnotemark[\value{footnote}].
It consists of tracked trajectories of $10$ basketball players (5 attackers, 5 defenders) during the $2016$ NBA season monitoring $1600$ matches. Each trajectory contains $50$ time steps sampled at $5~Hz$ with x, y, and z coordinates expressed in feet. $40$ time steps are used as observations and $10$ time steps for predictions. All trajectories are normalized and shifted to obtain zero-centred sequences to the middle of court.

\textbf{Intersection Drone Dataset (inD)}~(\cite{ind}). It captures four different German intersections from a bird's-eye-view perspective and contains more than $11000$ trajectories of various road users (e.g., pedestrians, cars, cyclists) saved in $33$ recordings. Data is collected at $25~Hz$ using a drone.

\textbf{TrajNet++}~(\cite{trajnetpp}). It is a large scale interaction-centric trajectory prediction benchmark composed of a real-world dataset and a synthetic dataset. The real-world dataset contains selected trajectories of different datasets (ETH~\cite{pellegrini}, UCY~\cite{ucy}, WildTrack~\cite{wt}, L-CAS~\cite{lcas} and CFF~\cite{cff}). This benchmarks defines a \textit{primary} pedestrian per scene and his/her categorization into four different types: static, linear, interacting and non-interacting.

\subsection{Metrics}

\textbf{TopK Average Displacement Error} (TopK ADE): Average Euclidean distance over all estimated points and ground-truth positions of a trajectory as proposed in \cite{pellegrini}:
\begin{equation}
    ADE = \sum_{i=1}^{\mathcal{P}}\sum_{t=T_{obs}+1}^{T_{pred}}\frac{\sqrt{(\hat{x}^i_t-{x^i_t})^2+(\hat{y}^i_t-{y^i_t})^2}}{T_{pred}\cdot\mathcal{P}};
\end{equation}

\textbf{TopK Final Displacement Error} (TopK FDE): Average Euclidean distance between predicted and ground-truth final destinations:
\begin{equation}
     FDE =\sum_{j=1}^{\mathcal{P}}\frac{\sqrt{(\hat{x}_{T_{pred}}^j-{x}_{T_{pred}}^j)^2+(\hat{y}_{T_{pred}}^j-{y}_{T_{pred}}^j)^2}}{{\mathcal{P}}}.
\end{equation}
$\mathcal{P}$ represents the number of pedestrians and $T_{pred}$ is the predicted time horizon. The above metrics are evaluated using the top-k (or best-of-N) i.e., we sample N trajectories and consider the ADE and FDE of the lowest-error trajectory.

\textbf{Average Log-Likelihood} (Avg NLL):
Average Log-Likelihood of ground truth trajectories over the predicted time horizon considering a distribution fitted with N output predictions. We compute this metric as in (\cite{trajnetpp}).

\textbf{TopK Collisions}: Similarly to (\cite{trajnetpp}), we consider two types of collisions, Col-I and Col-II, measuring the collisions of a pedestrian w.r.t his/her neighbours considering a fixed neighbourhood. Col-I (or prediction collision) uses the neighbours' predicted trajectories to check a collision, while Col-II relies on their ground-truth annotations. Nevertheless, since we use these metrics in a multi-modal context, we consider predictions with the lowest ADE (TopK ADE) for both primary and neighbours pedestrians. We report the percentage of collisions averaged over all test scenes.

\subsection{Quantitative Results}
\textbf{ETH-UCY.}
We evaluate our model using different versions of ETH-UCY datasets since multiple data and protocols are available for these scenes.
Quantitative results are reported in Table~\ref{tab:ethucy_tab}. We indicate with AC-VRNN our full model including the hidden state refinement process and belief maps and with A-VRNN our model without belief maps. Firstly, we consider a leave-one-out training protocol (A) as in S-GAN~(\cite{sgan}).
Our model outperforms all baselines on \emph{Eth} (FDE) and \emph{Zara2} (TopK ADE and TopK FDE with $K=20$) scenes and exhibits the best values on average metrics. AC-VRNN significantly outperforms A-VRNN suggesting the beneficial effect of belief maps conditioning.
For the remaining scenes, slightly worse performance of AC-VRNN could be ascribed to the leave-one-out protocol since training belief maps may not entirely comply with test scenes increasing uncertainty for future predictions.
SR-LSTM~(\cite{srlstm}) defines different \emph{Eth} annotations considering $6$ frames at $0.4s$ instead of $10$ frames due to a frame rate issue of original annotations, affecting each cross-validation fold (B). In this case, our model outperforms SR-LSTM baseline or achieve comparable results on all scenes for both metrics.
Finally, S-Ways~(\cite{socialways}) does not use a leave-one-out protocol. Each dataset is split into 5 subsets, using 4 subsets for training and the remaining ones for testing purpose (C). We achieve better performance on TopK ADE and slightly worse performance on TopK FDE.
Without the leave-one-out protocol, AC-VRNN significantly outperforms A-VRNN on FDE suggesting the beneficial effect of belief maps conditioning.

\textbf{Stanford Drone Dataset.}
To consider more complex urban scenarios, we test our model also on Stanford Drone Dataset. We compare our results with S-GAN-P~(\cite{sgan}) and STGAT~(\cite{stgat}).
As shown in Table~\ref{tab:sdd_tab}, AC-VRNN outperforms A-VRNN version and both selected baselines.
With more complex trajectories and scene topologies, our attentive module is able to better capture interactions among pedestrians and belief maps help to avoid incorrect behaviors following the prior distribution of displacements in the monitored scene.



\begin{table*}[!t]
\centering
\scalebox{0.8}{
    \begin{tabular}{lcl|ccccc|c}
    \hline \toprule \noalign{\smallskip}
    &&Method       & \textbf{ETH} & \textbf{HOTEL} & \textbf{UNIV} & \textbf{ZARA1} & \textbf{ZARA2} & \textbf{AVG} \\ 
    \noalign{\smallskip} \midrule
    \multirow{9}{*}{(A)}
    &&S-LSTM (\cite{slstm})       &  1.09/2.35  &  0.79/1.76  &  0.67/1.40  &  0.47/1.00  &  0.56/1.17  &  0.72/1.54   \\
    &&S-GAN-P (\cite{sgan})  &  0.87/1.62  & 0.67/1.37  &  0.76/1.52  &  0.35/0.68  &  0.42/0.84  &  0.61/1.21    \\
    &&S-GAN (\cite{sgan}) &  0.81/1.52  & 0.72/1.61  &  0.60/1.26  &  0.34/0.69  &  0.42/0.84  &  0.58/1.18 \\
    && Trajectron (\cite{trajectron}) & \textbf{0.59}/1.14 & 0.35/0.66 & 0.54/1.13 & 0.43/0.83 & 0.43/0.85 & 0.56/1.14 \\
    &&SoPhie (\cite{sophie})      &  0.70/1.43  &  0.76/1.67  &  0.54/1.24  &  \textbf{0.30}/0.63  &  0.38/0.78  & 0.54/1.15    \\ 
    &&Social-BiGAT (\cite{social-bigat}) &  0.69/1.29  &  0.49/1.01  &  0.55/1.32  &  \textbf{0.30}/\textbf{0.62}  &  0.36/0.75  & 0.48/1.00    \\ 
    &&Next (\cite{peekingintothefuture})         & 0.73/1.65   &  \textbf{0.30}/0.59  &  0.60/1.27  &  0.38/0.81  & 0.31/0.68  &  0.46/1.00    \\ 
    &&STGAT (\cite{stgat}) &  0.78/1.60  &  \textbf{0.30}/\textbf{0.54}  & \textbf{0.51}/\textbf{1.08}  & 0.33/0.72  & 0.29/0.63  & 0.44/0.91    \\
    &&A-VRNN (Ours)  &  0.73/1.45 &	0.34/0.65 &	0.53/1.14 &	0.33/0.69 &	\textbf{0.26}/\textbf{0.54} &		0.44/0.89   \\ 
    &&AC-VRNN (Ours) & 0.61/\textbf{1.09} &  \textbf{0.30}/0.55 &	0.58/1.22 &	0.34/0.68 &	0.28/0.59 &	\textbf{0.42}/\textbf{0.83} \\ \midrule
    \multirow{2}{*}{(B)}
    &&SR-LSTM (\cite{srlstm})     &  0.63/1.25  &  \textbf{0.37}/\textbf{0.74}  &  \textbf{0.51}/\textbf{1.10}  &  0.41/0.90  &  0.32/0.70  &  0.45/0.94 \\ 
    &&A-VRNN (Ours)  & \textbf{0.60}/\textbf{1.18}  &	\textbf{0.37}/\textbf{0.74}  & 0.55/1.20 &	\textbf{0.39}/\textbf{0.83} &	\textbf{0.27}/\textbf{0.59} &	\textbf{0.44}/\textbf{0.91}    \\ \midrule
    \multirow{3}{*}{(C)}
    &&S-Ways (\cite{socialways}) &  \textbf{0.39}/\textbf{0.64}  &  0.39/0.66  &  \textbf{0.55}/\textbf{1.31}  &  0.44/\textbf{0.64}  &  0.51/0.92  &  0.46/\textbf{0.83} \\ 
    &&A-VRNN (Ours) & 0.60/1.24 & 0.22/0.45 & 0.61/1.34 & 0.46/1.06 & \textbf{0.30}/\textbf{0.67} & \textbf{0.44}/0.95 \\ 
    &&AC-VRNN (Ours) & 0.55/1.06 & \textbf{0.18}/\textbf{0.26} & 0.76/1.59 & \textbf{0.37}/0.72 & 0.33/0.70 & \textbf{0.44}/0.87 \\ \bottomrule
    \end{tabular}
    }
    \caption{Quantitative results of considered methods for ETH and UCY datasets. We report Average Displacement Error (ADE) and Final Displacement Error for unimodal methods and TopK ADE and TopK FDE (with $K=20$) for multi-modal ones. The results were obtained for $t_{obs}=8$ and $t_{pred}=12$ (in meters). The first block of experiments regards the using of data employed by S-GAN and STGAT models; the second one uses the SR-LSTM version of data while the last experiments are trained with the S-Ways protocol. On average, our model outperforms several methods showing a slightly worse FDE error when the S-Ways protocol is employed. No belief maps appear necessary for SR-LSTM data version.}
    \label{tab:ethucy_tab}
\end{table*}

\begin{table}[!t]
 \centering
 \resizebox{\columnwidth}{!}{
     \begin{tabular}{l|ccccc}
     \hline \toprule\noalign{\smallskip}
     Method   & & & \textbf{SDD} \\
     & TopK ADE ($\downarrow$) & TopK FDE ($\downarrow$) & Avg NLL ($\uparrow$) & Col-I ($\downarrow$) & Col-II ($\downarrow$) \\
     \noalign{\smallskip} \midrule
     S-GAN-P (\cite{sgan})  & $0.65$ & $1.26$ & $-3.79$ & $0.00$ & $0.33$ \\
     STGAT (\cite{stgat})   & $0.57 $ & $1.09 $ & $-2.70$ & $0.00$ & $0.40$\\
     DAG-Net (\cite{dagnet}) & $0.54 $ & $1.07$ & $-2.54$  & $0.49$ & $0.25$\\ 
     A-VRNN (Ours)  & $0.55$ & $0.98$ & $-1.11$ & $\textbf{0.00}$ & $\textbf{0.11}$\\
     AC-VRNN (Ours) & $\textbf{0.51}$ & $\textbf{0.90} $ & $\textbf{-0.18} $ & $0.16$ & $0.22$ \\ \bottomrule 
     \end{tabular}}
     \caption{Results for $t_{obs}=8$ and $t_{pred}=12$ on Stanford Drone Dataset (in meters). AC-VRNN significantly reduces TopK ADE and TopK FDE error metrics. Average NLL is the best one among all approaches while collision errors are below 1\% for all methods.}
     \label{tab:sdd_tab}
\end{table}

\begin{table}[!t]

 \resizebox{\columnwidth}{!}{
 \centering
     \begin{tabular}{l|c|ccccc}
     \hline \toprule\noalign{\smallskip}
     Team & Method  & & & \textbf{STATS SportVU NBA} \\
     & & TopK ADE ($\downarrow$) & TopK FDE ($\downarrow$) & Avg NLL ($\uparrow$) & Col-I ($\downarrow$) & Col-II ($\downarrow$) \\
     \noalign{\smallskip} \midrule
     & STGAT (\cite{stgat})   & $9.94 $ & $15.80 $ & $-8.65$ & $0.21$ & $0.32$ \\
     ATK & Weak-Supervision (\cite{weeksup}) & $9.47 $ & $16.98 $ & $\textbf{-6.29}$ & $0.57$ & $0.20$\\
     & A-VRNN (Ours)  & $\textbf{9.32} $ & $ \textbf{14.91} $ &  $-7.60$ & $\textbf{0.09}$ & $\textbf{0.18}$  \\ \midrule
     & STGAT (\cite{stgat})   & $7.26 $ & $11.28 $ & $-7.88$ & $0.20$ & $\textbf{0.27}$ \\
      DEF & Weak-Supervision (\cite{weeksup}) & $7.05 $ & $10.56 $ & $\textbf{-5.69}$ & $0.70$ & $0.57$ \\
     & A-VRNN (Ours)  & $\textbf{7.01} $ & $\textbf{10.16} $  & $-6.70$ & $\textbf{0.13}$ & $0.43$ \\ \bottomrule 
     \end{tabular}}
     \caption{Results for $t_{obs}=10$ and $t_{pred}=40$ in feet on STATS SportVU NBA dataset.}
     \label{tab:basket_tab}
\end{table}

\begin{table}[!t]
 \resizebox{\columnwidth}{!}{
 \centering
     \begin{tabular}{l|cccccc}
     \hline \toprule\noalign{\smallskip}
     Method   & & & \textbf{inD} \\
     & TopK ADE ($\downarrow$) & TopK FDE ($\downarrow$) & Avg NLL ($\uparrow$) & Col-I ($\downarrow$) & Col-II ($\downarrow$) \\
     \noalign{\smallskip} \midrule
     S-GAN (\cite{sgan})  & $0.48$ & $0.99$ & $-1.84$ & $\textbf{0.51}$ & $0.55$ \\
     STGAT (\cite{stgat}) & $0.48$ & $1.00$ &  $-1.55$ & $0.60$ & $0.58$ \\
     A-VRNN (Ours) &  $0.45$ & $0.97 $ & $-1.69$ & $0.61$ & $\textbf{0.52}$ \\
     AC-VRNN (Ours) &  $\textbf{0.42}$ & $\textbf{0.80} $ & $\textbf{-0.29}$ & $0.78$ & $0.61$ \\\bottomrule
     \end{tabular}}
     \caption{Results for $t_{obs}=8$ and $t_{pred}=12$ in meters on inD dataset.}
     \label{tab:ind_tab}
\end{table}

\begin{table}[!t]
 \resizebox{\columnwidth}{!}{
 \centering
     \begin{tabular}{l|ccc}
     \hline \toprule\noalign{\smallskip}
     Method & \hspace{4cm} \textbf{TrajNet++} \\
     & ADE/TopK ADE ($\downarrow$) & FDE/TopK FDE ($\downarrow$) \\
     \noalign{\smallskip} \midrule
     S-LSTM (\cite{slstm}) & $0.55$ & $1.18$  \\
     S-ATT (\cite{socialattention}) & $0.56$ & $1.22$  \\ 
     S-GAN (\cite{sgan})  & $\textbf{0.51}$ & $\textbf{1.09}$ \\ 
     D-LSTM(\cite{trajnetpp}) & $0.57$ & $1.23$ \\
     AC-VRNN (Ours) & $0.57$ & $\underline{1.17}$ \\ \bottomrule
     \end{tabular}}
     \caption{Results for $t_{obs}=9$ and $t_{pred}=12$ in meters on TrajNet++. For unimodal methods ADE and FDE metrics are reported while for multimodal ones we reported the TopK ADE and TopK FDE metrics with $K=3$.}
     \label{tab:trajnet++_tab}
\end{table}

\textbf{STATS SportVU NBA.}
Additionally, we test our model using basketball players trajectories whose dynamics are clearly different from ones exhibited by pedestrians in urban scenes. As reported in Table~\ref{tab:basket_tab}, our A-VRNN reduces TopK ADE and TopK FDE metrics on both offensive and defencive players trajectories compared to STGAT~\cite{stgat} and Weak-Supervision~\cite{weeksup}. Avg NLL are similar for all methods, 
whereas collision errors given by A-VRNN are mainly smaller than the errors generated by competitive approaches.
In this case, belief maps cannot properly steer future positions since basketball courts do not have obstacles and never-crossed areas. Moreover, basketball players do not typically follow a collective behaviour.

\textbf{Intersection Drone Dataset.}
On InD dataset, we adopt the same evaluation protocol used for Stanford Drone Dataset considering, for each scene, $70\%$ of data as training, $10\%$ as validation while the remaining part for testing. We retain only pedestrians' trajectories and downsample each scene to obtain $20$ time steps in $8$ s. In Table~\ref{tab:ind_tab} we compare our model to S-GAN~(\cite{sgan}) and STGAT~(\cite{stgat}). AC-VRNN overcomes all the competitive methods on TopK ADE and TopK FDE and Avg NLL. S-GAN gives a smaller Col-I error with respect to AC-VRNN and S-GAN, while A-VRNN shows a smaller Col-II error.

\textbf{TrajNet++.}
Finally, we test our model on TrajNet++~(\cite{trajnetpp}) real-world dataset. The results are reported in Table~\ref{tab:trajnet++_tab} where ADE and FDE metrics are used for unimodal methods and TopK ADE and TopK FDE (with $K=3$) metrics for multimodal ones. We find that our model reaches competitive performance with respect to other approaches, especially for TopK FDE. Our results are obtained by submitting the results to the evaluation server averaging the results on different types of scene considering only the real dataset. We compare AC-VRNN against published competitive approaches as competing with a lead-board that is updated every day is out of the scope of this quantitative analysis. Other methods results are reported from (\cite{trajnetpp,snce}). Since the Avg NLL for competitive methods is missing, we do not report this metric in Table~\ref{tab:trajnet++_tab}. However, our method attains an Avg NLL of $-8.33$.

\subsection{Ablation Experiments}
We also present an ablation study to show the contribution of different components of our model on the prediction task. In the following, we detail each component and report quantitative results in Table~\ref{tab:ablation} and Table~\ref{tab:abl_neigh}.

\textbf{Vanilla Variational Recurrent Network.}
We investigate the ability of Vanilla VRNNs to predict accurate trajectories on ETH, UCY and SDD datasets. This model does not consider any human interactions or prior scene knowledge. ETH scenes appear mainly affected by the lack of additional information while UCY scenes attain comparable results to our AC-VRNN model, especially for TopK ADE metric. Such a result highlights the importance of trajectory forecasting task to go beyond a time-series problem and the need of including contextual information about the scene, such as human interactions or experience gained in similar contexts.

\textbf{Hidden State Refinement with Graph Convolutional Neural Network.}
This experiment models interactions with a hidden state refinement based on a standard Graph Convolutional Networks (GCN)
. The model has worse performance compared to AC-VRNN and Vanilla VRNN models on ETH and UCY datasets while obtains comparable results to AC-VRNN on SDD dataset. The experiments suggest that, for complex contexts, attention mechanisms are able to capture more useful information in order to model interactions among pedestrians compared to simple scenarios where interactions may be reduced.

\textbf{AC-VRNN without KLD Loss on Belief Maps.}
To demonstrate the importance of KL-divergence loss on belief maps, we train our model without this term yet still conditioning the model on them. We obtain the worst results on all datasets proving that the network is not able to integrate belief maps information conditioning only VAE components. KL-divergence allows the network to generate displacement distributions similar to the ground-truth ones and to follow prior knowledge about local behaviors.

\textbf{Adjacency Matrix.}
 We also evaluate our model using different kinds of adjacency matrices to corroborate the use of the similarity one.
 We consider an \textit{all-1} adjacency matrix where edges are equally weighted and all pedestrians in the scene are connected. This model attains good performance but slightly worse than the ones obtained with a similarity matrix on both ETH/UCY and SDD, proving that assuming the same importance for all involved agents negatively affects the results.
 $k$-NN matrix only considers nearby pedestrians. The neighbourhood is computed by sorting mutual distances between each pedestrian, retaining only the first $k$ nearest neighbours (with $k=3$), defined as a set $S_{i}$. Each element is set to $1$ if $a_{i,j} \in S_{i}$, to $0$ otherwise.
 $k$-NN matrix obtains quite the worst results on ETH and UCY datasets and performs poorly on SDD dataset. This experiment demonstrates that a small neighbourhood is not able to capture interactions in large scenes where pedestrians show mutual influences also at long distances.

\textbf{Belief Maps Dimension.}
Since belief maps define the probability that a pedestrian in a cell will move towards another one, it is important to consider a proper cell dimension. If we consider a fine-grained grid ($L=3$), we could discard information about pedestrians whose displacement is greater than the defined one. Likewise, if we consider a course-grained grid ($L=9$), outermost cells may not be properly filled. To select the best value of the parameter $L$, we test our model using different cell dimensions and found that $L=5$ is the best choice for our datasets.
\begin{table}[!t]
 \centering
 \resizebox{\columnwidth}{!}{
 \setlength{\tabcolsep}{1.5mm}
   \renewcommand{\arraystretch}{1.1}
     \begin{tabular}{l|ccccc|c}
     \hline \toprule \noalign{\smallskip} 
     Method  & \textbf{ETH} & \textbf{HOTEL} & \textbf{UNIV} & \textbf{ZARA1} & \textbf{ZARA2} & \textbf{AVG} \\
     \noalign{\smallskip} \midrule
     Vanilla VRNN   & 0.79/1.61	& 0.46/0.94 & 0.55/1.20 & 0.34/0.75 & 0.26/0.58 & 0.48/1.02 \\
     GCN-VRNN & 0.81/1.58 & 0.41/0.85 & 0.59/1.31 & 0.38/0.84 & 0.41/0.96 & 0.52/1.11 \\
     AC-VRNN w/o KLD & 0.73/1.41 & 0.52/1.07 & 0.64/1.36 & 0.43/0.89 & 0.39/0.83 & 0.54/1.11 \\
     All-1 ADJ Matrix & 0.77/1.52 & 0.37/0.73 & 0.55/1.19 & 0.34/0.75 & 0.26/0.58 & 0.46/0.95 \\
     kNN ADJ Matrix & 0.76/1.54 & 0.47/0.99 & 0.57/1.26 & 0.42/0.95 & 0.26/0.58 & 0.50/1.01 \\ \midrule
     A-VRNN (Ours) &  0.73/1.45 &	0.34/0.65 &	\textbf{0.53}/\textbf{1.14} &	\textbf{0.33}/0.69 &	\textbf{0.26}/\textbf{0.54} &	0.44/0.89 \\
     AC-VRNN (Ours) & \textbf{0.61}/\textbf{1.09} &  \textbf{0.30}/\textbf{0.55} &	0.58/1.22 &	0.34/\textbf{0.68} &	0.28/0.59 &	\textbf{0.42}/\textbf{0.83} \\  \bottomrule
     \end{tabular}
     }
     \caption{Ablation experiments showing TopK ADE and TopK FDE for $t_{obs}=8$ and $t_{pred}=12$ in meters on ETH, UCY and SDD datasets. AVG column reports average results for ETH and UCY datasets.}
     \label{tab:ablation}
\end{table}

\begin{table}[!t]
\centering
\scalebox{0.8}{
    \begin{tabular}{l|cc}
    \hline \toprule \noalign{\smallskip} 
     Method & \hspace{3cm} \textbf{SDD} \\
     & TopK ADE ($\downarrow$) & TopK FDE ($\downarrow$) \\
     \noalign{\smallskip} \midrule
     Vanilla VRNN & $0.56 $ & $1.15 $ \\
     GCN-VRNN & $0.53 $ & $1.05 $ \\
     AC-VRNN w/o KLD & $0.60 $ & $1.11 $ \\
     All-1 ADJ Matrix & $0.57 $ & $1.11 $ \\
     kNN ADJ Matrix & $0.73 $ & $1.43 $ \\
     A-VRNN & $0.56 $ & $1.14 $ \\
     AC-VRNN ($L=3$) & $ 0.67 $ & $1.31$ \\
     AC-VRNN ($L=5$) & $\textbf{0.51} $ & $\textbf{0.92}$ \\
     AC-VRNN ($L=7$) & $0.68 $ & $1.33$ \\ \bottomrule
    \end{tabular}
    }
    \caption{Ablation experiments showing TopK ADE and TopK FDE for $t_{obs}=8$ and $t_{pred}=12$ for SDD dataset.}
    \label{tab:abl_neigh}
\end{table}

\textbf{Hidden State Initialization.}
The hidden state initialization has a strong impact on the RNN training process. We experiment with three different initialization approaches:
\begin{itemize}
    \item \emph{Zero initialization}: a simple zero-tensor initialization.
    \item \emph{Learned initialization}: a linear layer is trained to learn an optimal initialization. 
    \item \emph{Absolute coordinate initialization}: the tensor is initialized with the first absolute coordinates to provide spatial information to the learning process that is based on displacements generation.
\end{itemize}
We experimentally notice that the \emph{absolute coordinate initialization} has a significant impact on the recurrent process leading to a performance improvement on ETH/UCY dataset and on SDD, while on STATS SportVU NBA InD and TrajNet++ the \emph{zero initialization} is preferable.

\textbf{Block Irregular Adjacency Matrix.}
AC-VRNN is based on a single Variational Recurrent Neural Network with shared parameters. To jointly compute a unique adjacency matrix for each time step, we build a block matrix where each block contains the matrix corresponding to a single scene, randomly chosen from the training dataset. Blocks can have different dimensions since a variable number of agents may be present in the scene.

\begin{figure*}
\centering
\includegraphics[height=4.6cm, width=5.8cm]{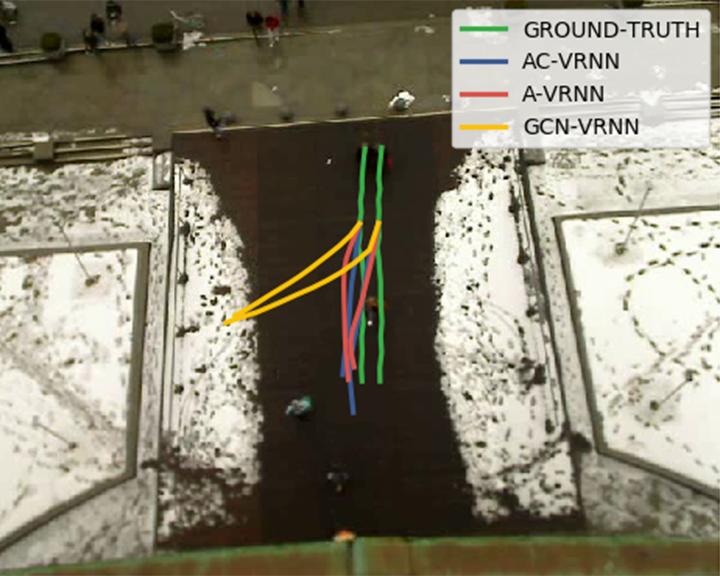} 
\includegraphics[height=4.6cm, width=5.8cm]{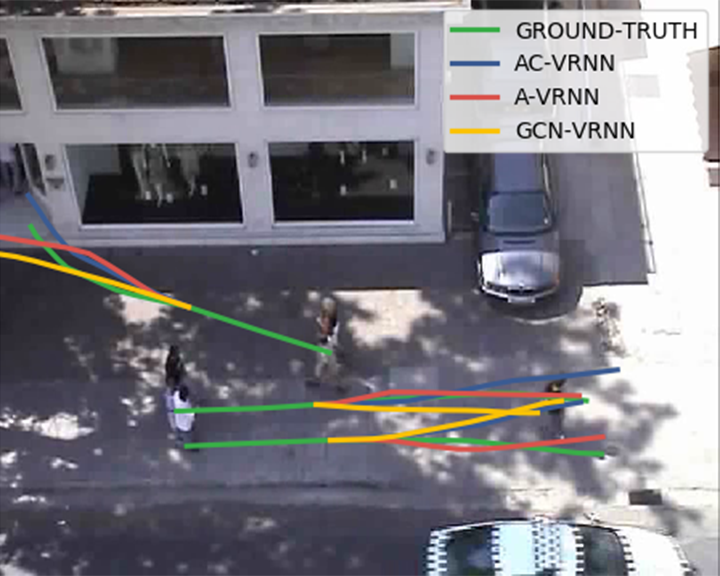}
\includegraphics[height=4.6cm, width=5.8cm]{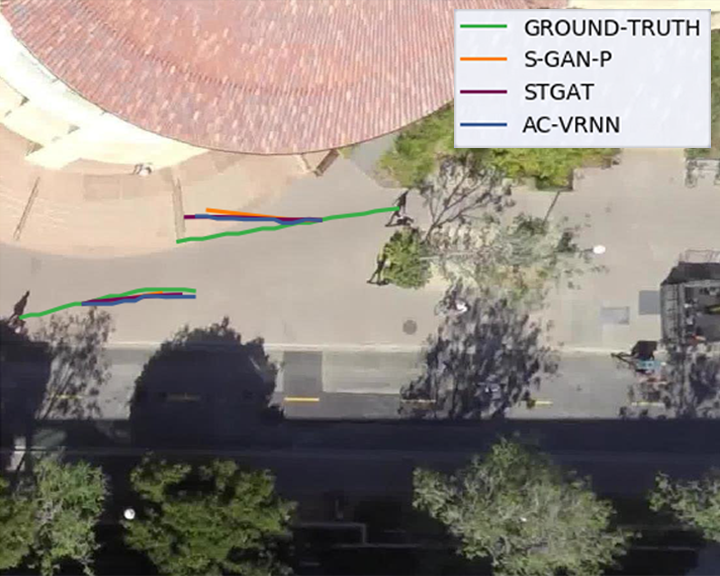} \\
\includegraphics[height=4.6cm, width=5.8cm]{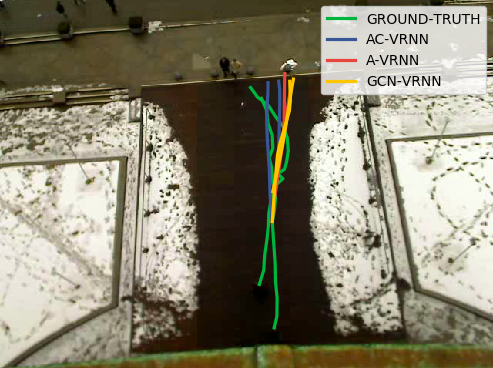}
\includegraphics[height=4.6cm, width=5.8cm]{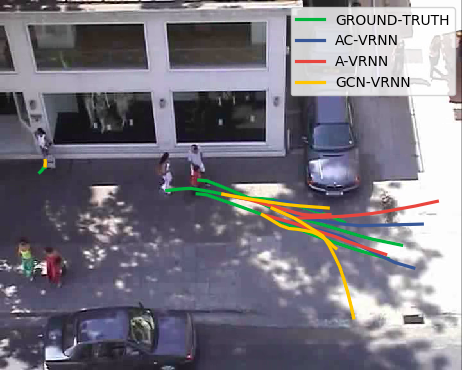}
\includegraphics[height=4.6cm, width=5.8cm]{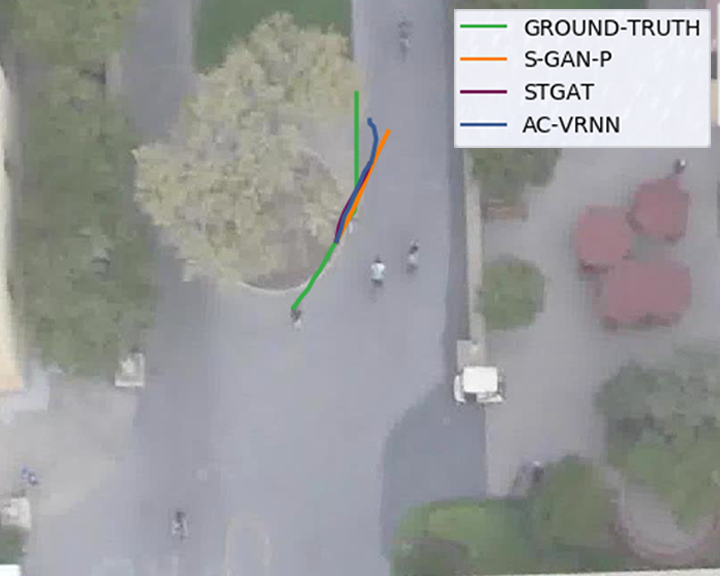}

\caption{Illustration of predicted trajectories using AC-VRNN, baselines and competitive methods on \emph{Eth} (left) and \emph{Zara1} (middle) scenes of ETH and UCY datasets and \emph{gates\_0} and \emph{deathCircle\_1} of SDD (right).}
\label{fig:imgs}
\end{figure*}

\subsection{Qualitative Results}
Figure~\ref{fig:imgs} presents some qualitative experiments, comparing our model with baselines and competitive methods. On \emph{Eth}, GCN-VRNN, based on a Graph Convolutional Neural Network, generates trajectories that significantly drift from the ground-truth ones. On \emph{Zara1}, all considered models are able to follow correct paths, but AC-VRNN appears more able to predict complex trajectory such as the entrance into a building, following the collective agents' behaviour. For SDD, we randomly select two scenes and show our model samples against competitive methods. All methods predict plausible paths, but AC-VRNN generates more realistic trajectories in some cases, following the sidewalk rather than crossing the road diagonally.

\begin{figure*}[!t]
    \centering
    \includegraphics[height=3.5cm]{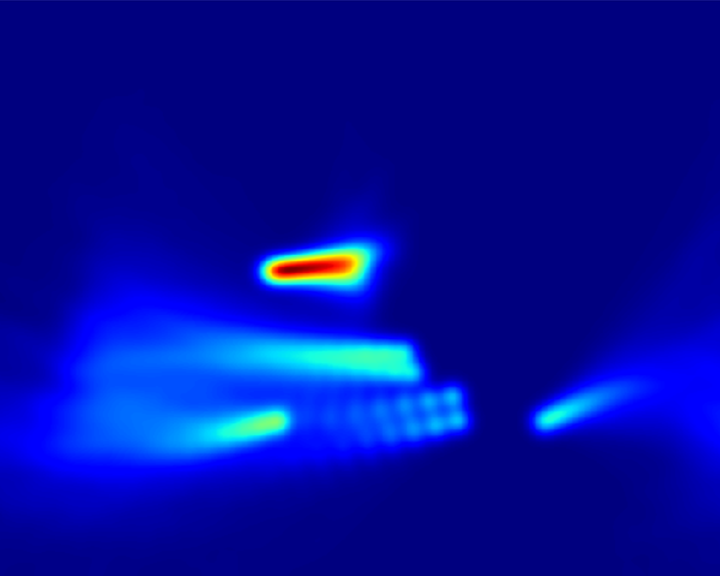} 
    \includegraphics[height=3.5cm]{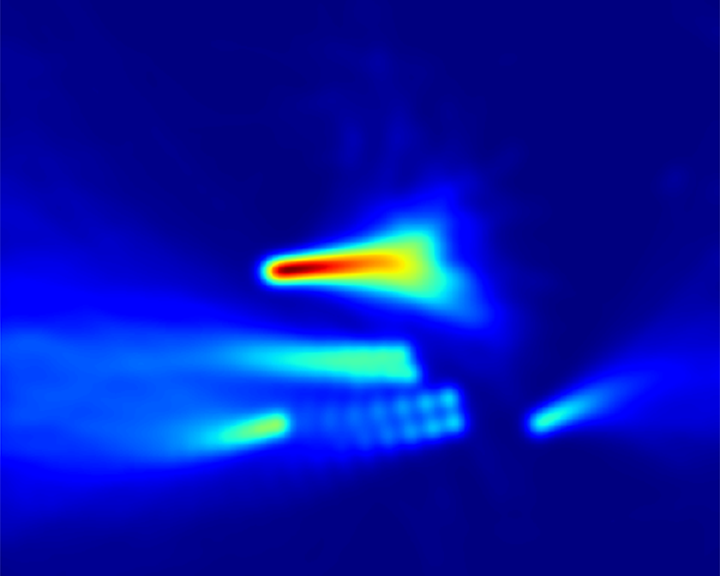}
    \includegraphics[height=3.5cm]{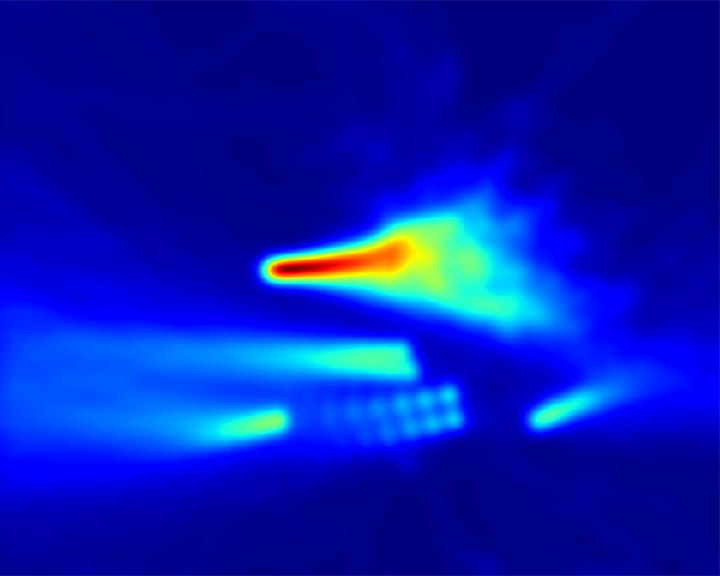}
    \includegraphics[height=3.5cm]{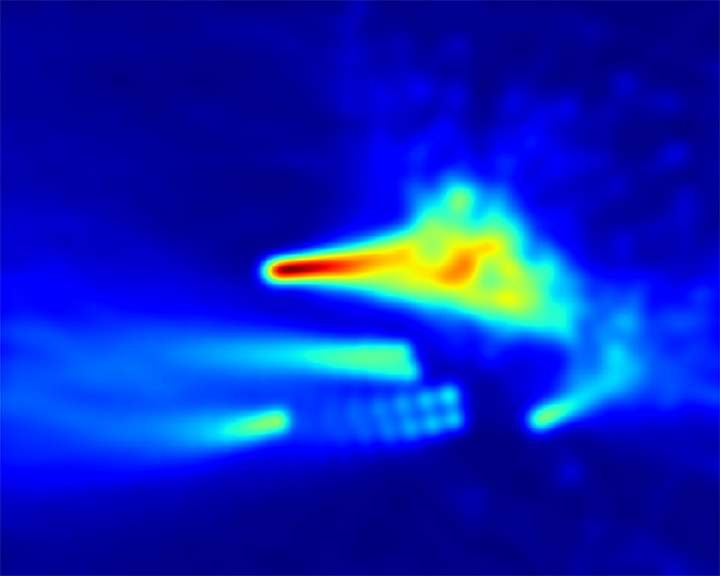}
    \caption{Heatmaps of the predictions probability distribution for long-term predictions. The predictions are made for $t_{obs}=8$ and $t_{pred}=20, 60, 120$ and  $200$, respectively (from left to right). We select \emph{Zara1} scene and observe that the trajectories are coherent with the scene topology.}
    \label{fig:long-term-predictions}
\end{figure*}

\textbf{Long-term predictions.}
Since AC-VRNN is a completely generative model, it is possible to generate an unlimited number of future positions as well as creating trajectories without any observations. This could be especially useful for applications that require sampling a large number of trajectories to simulate realistic motion dynamics as required by synthetic scenarios mimicking real-life situations.
Obviously, as the number of time steps increases, the predicted paths tend to drift from realistic ones, but our model qualitatively predicts plausible trajectories even after several time steps. To this end, we show in Figure~\ref{fig:long-term-predictions} some qualitative experiments considering up to $200$ time steps.

\begin{figure*}[!t]
    \centering
    \includegraphics[height=3.5cm, width=4.4cm]{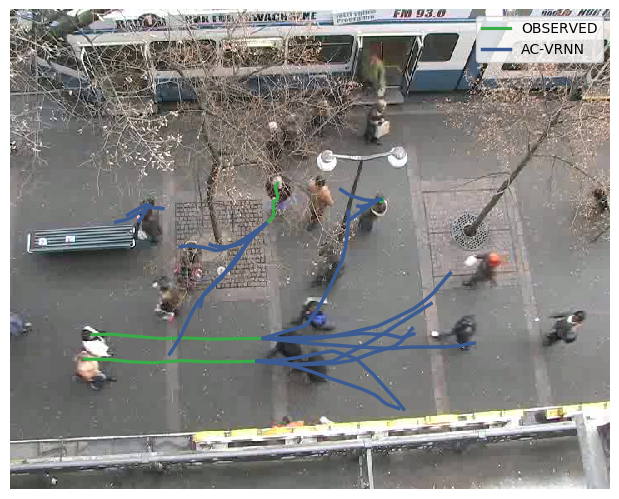}
    \includegraphics[height=3.5cm, width=4.4cm]{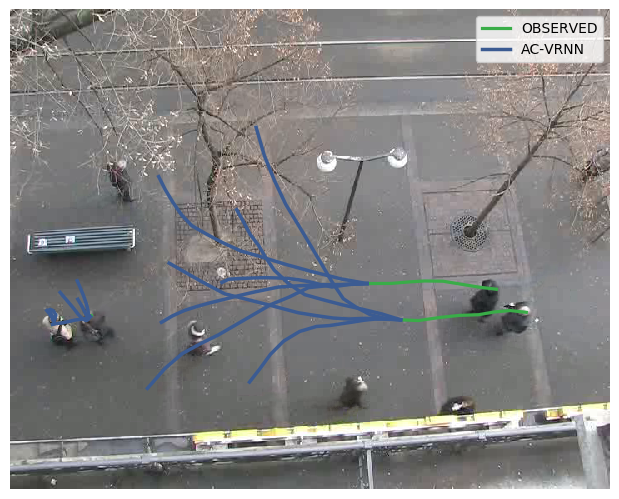}
    \includegraphics[height=3.5cm, width=4.4cm]{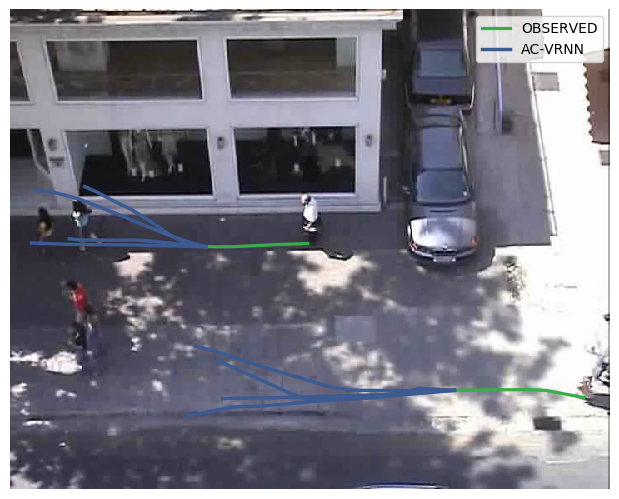} 
    \includegraphics[height=3.5cm, width=4.4cm]{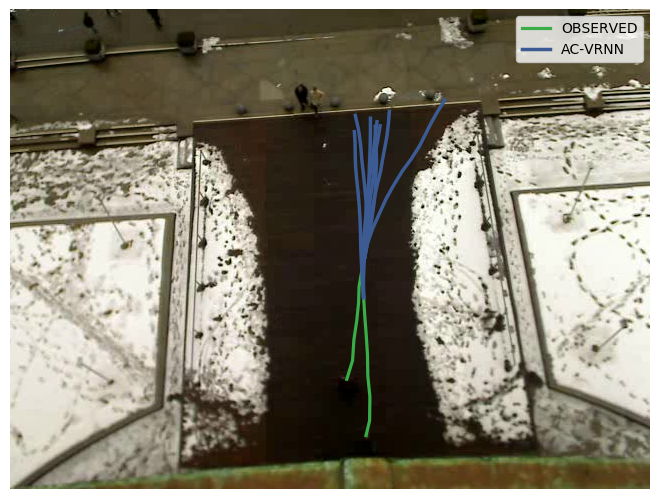} \\
    \includegraphics[height=3.5cm, width=4.4cm]{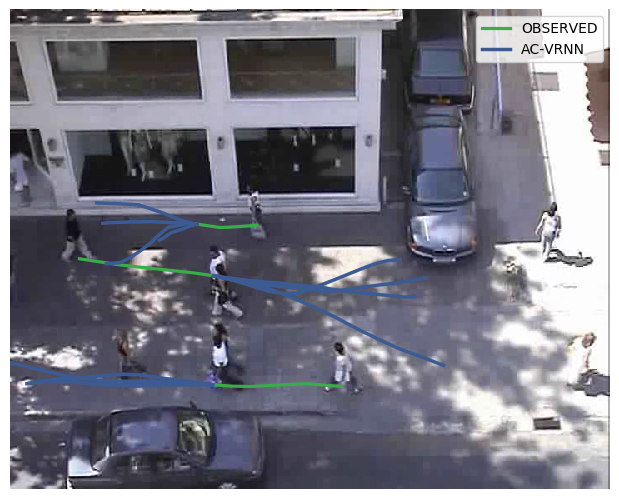}
    \includegraphics[height=3.5cm, width=4.4cm]{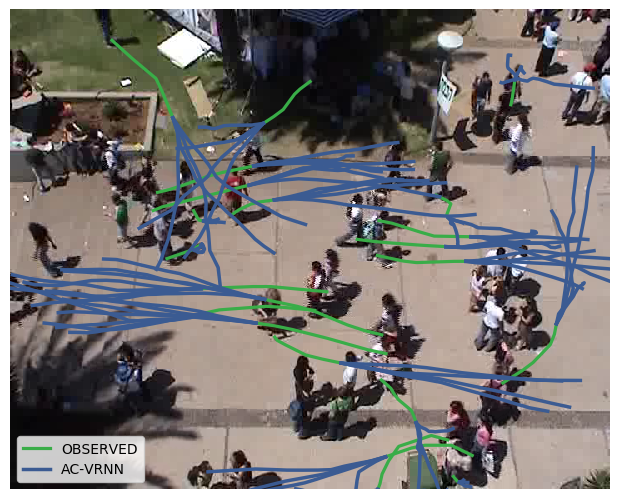}
    \includegraphics[height=3.5cm, width=4.4cm]{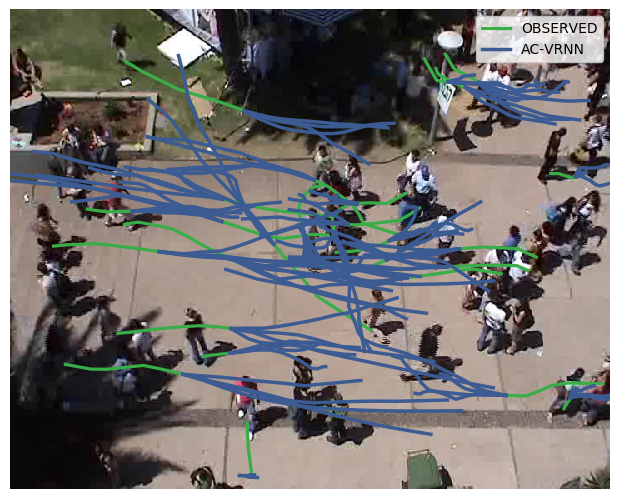}
    \includegraphics[height=3.5cm, width=4.4cm]{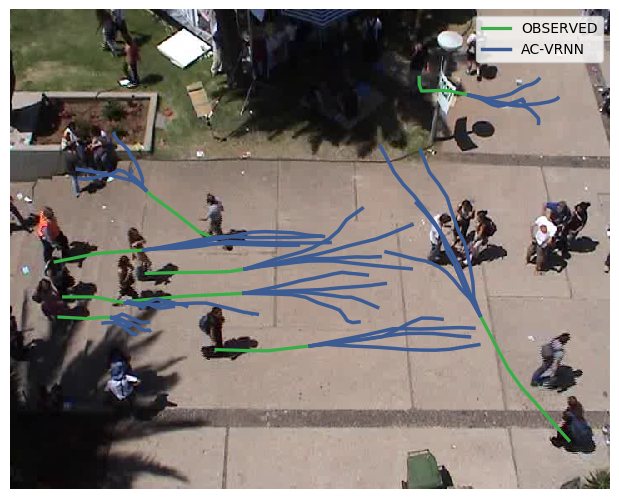}
    \caption{Multiple predictions of AC-VRNN trajectories to highlight the multi-modality nature of our model on ETH and UCY datasets.}
    \label{fig:cmpbas-multi-modal}
\end{figure*}

\begin{figure*}[!t]
    \centering
    \includegraphics[height=3.5cm]{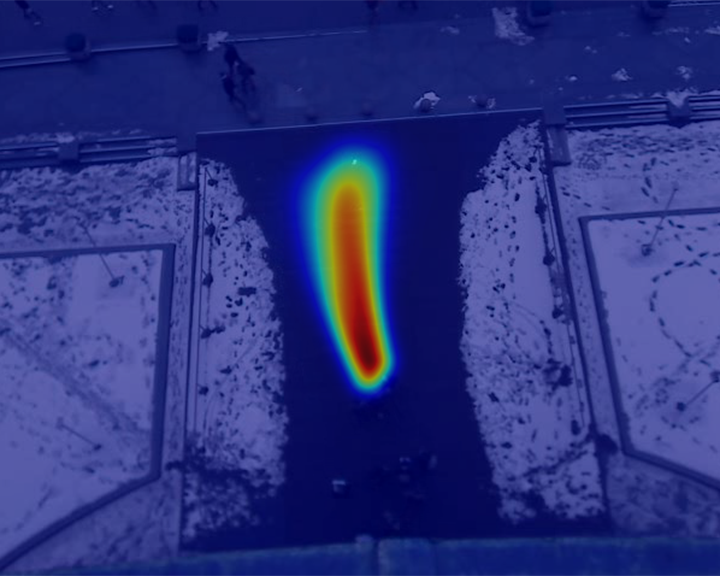}
    \includegraphics[height=3.5cm]{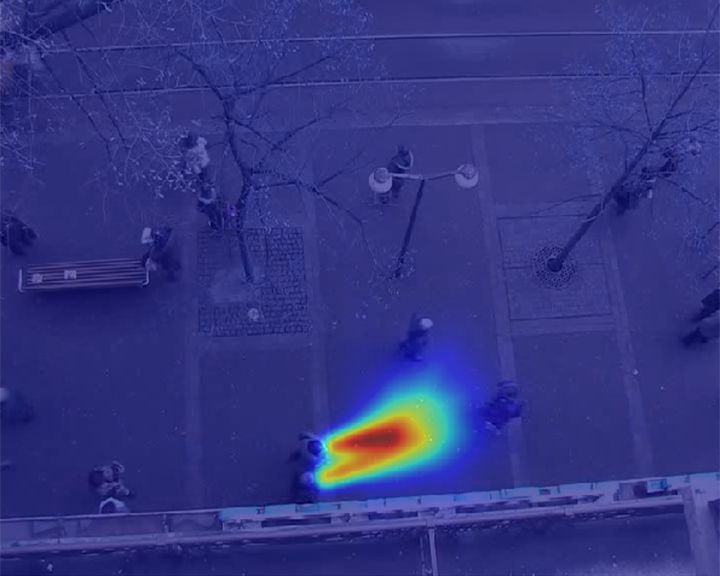} 
    \includegraphics[height=3.5cm]{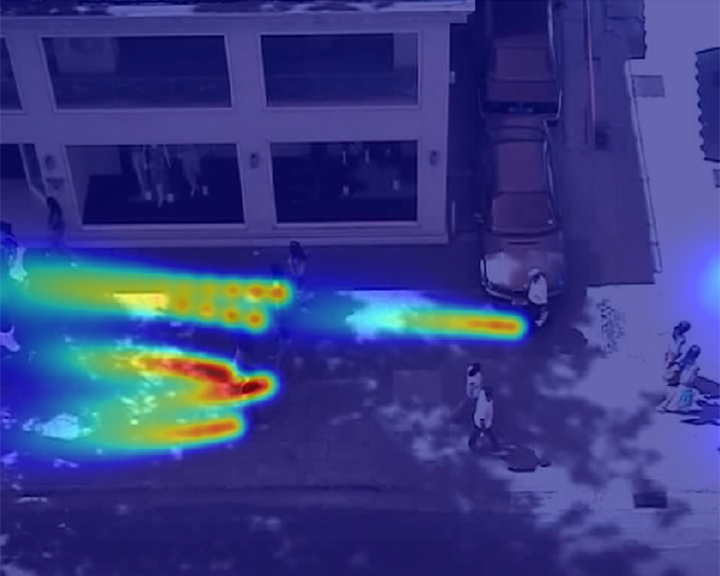} 
    \includegraphics[height=3.5cm]{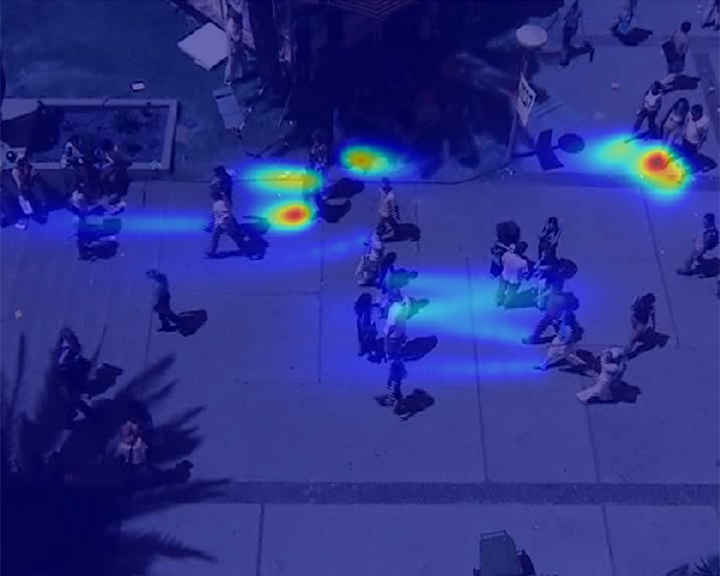} \\
    \includegraphics[height=3.5cm]{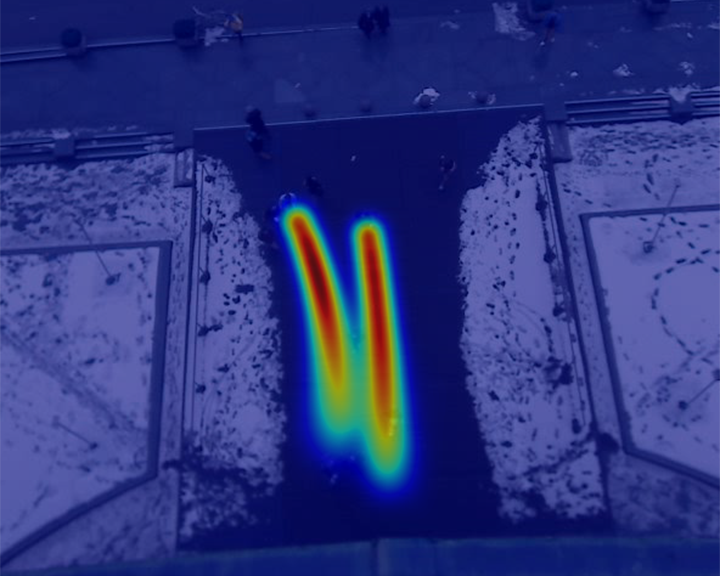} 
    \includegraphics[height=3.5cm]{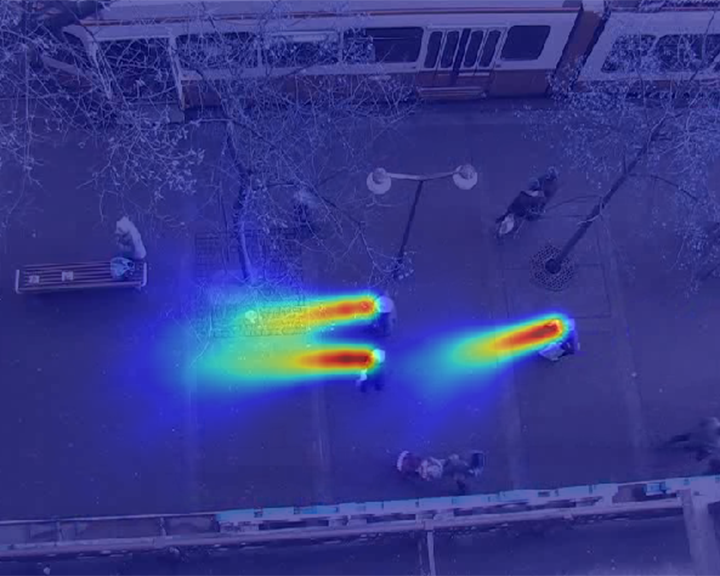}
    \includegraphics[height=3.5cm]{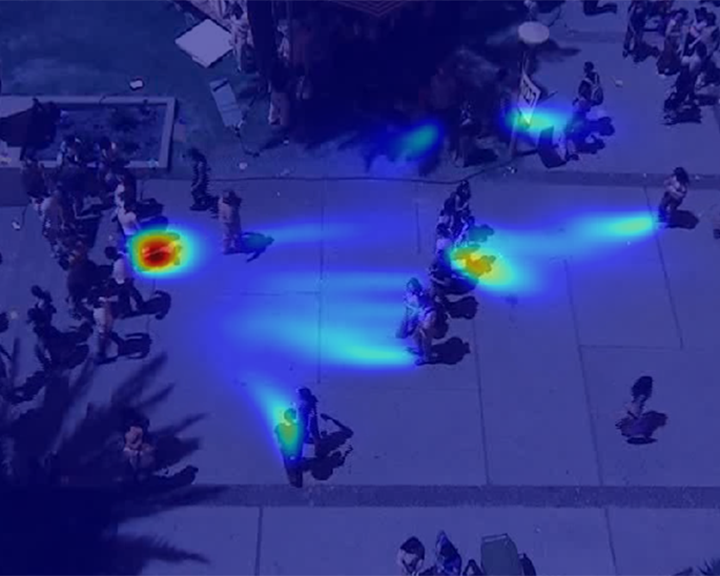}
    \includegraphics[height=3.5cm]{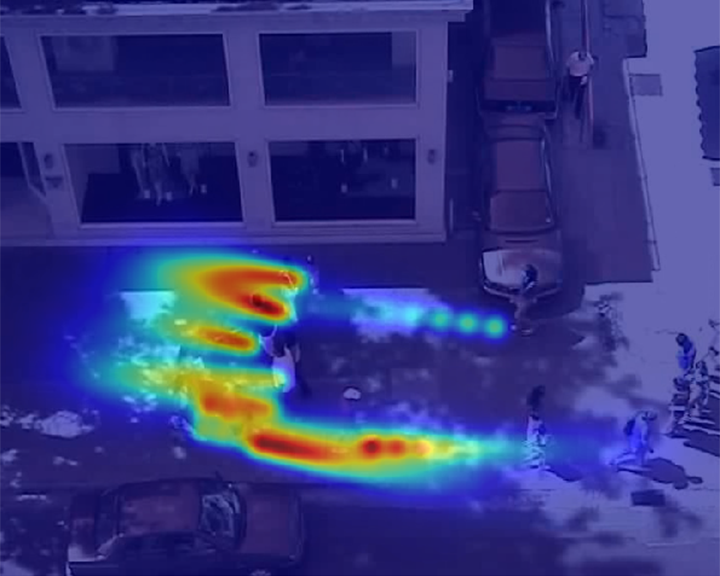} 
    \caption{Heatmaps representing probability distributions generated by our model for ETH and UCY datasets. }
    \label{fig:heatmaps}
\end{figure*}
\textbf{Multimodal predictions.}
Figure~\ref{fig:cmpbas-multi-modal} depicts other qualitative examples generated by AC-VRNN model showing multiple paths to demonstrate the ability of our model to predict multi-modal trajectories. Finally, Figure~\ref{fig:heatmaps} shows probability distributions of future paths. When interactions among pedestrians is limited or absent, our model correctly predicts continuous linear paths. By contrast, the increasing number of human interactions leads the predictions to simulate complex patterns.

\subsection{Implementation Details}
We train our model for $500$ epochs on ETH-UCY and SDD, for $300$ epochs on STATS SportVU NBA, for $300$ epochs on inD and for $25$ epochs on TrajNet++. Except for ETH/UCY, we re-train all competitive methods for the same number of epochs and report the best results after performing a hyperparameter search retaining the best model on the validation set. For ETH/UCY we report results from the original paper except for STGAT~(\cite{stgat}) that has been re-trained with the best hyperparameters proposed by the authors. We use gradient clipping set to $10$. For SGD optimizer we use a momentum of $0.9$. The RNN is a GRU with $1$ layer and hidden size equals to $64$. The attentive GNN has a hidden size of $8$ with $4$ attention heads. Each belief map during training is generated by sampling $100$ displacements. In Eq.~\ref{eq:loss}, $k$ is set to $100$ for all the datasets. Other hyperparameters that vary according to the dataset are reported in Table~\ref{tab:params}. In Table \ref{tab:backbone} an overall description of AC-VRNN architecture is reported.~\footnote{For a more detailed explanation see (\cite{gat}) and \url{https://github.com/Diego999/pyGAT}}

\textbf{Warm-up on VRNN KL-Divergence.} VRNN is trained with the ELBO loss that is composed of two terms: Negative Log-Likelihood and KL-Divergence. To correctly balance these two terms, we use a warm-up method that increases the weight in the range $[0,1]$ of the KL-Divergence up to $N$ epochs. After this learning period, we fix the KL weight to $1$. This technique favours the reconstruction error during the early epochs in order to firstly teach the network to generate correct samples and then to approach both \emph{encoder}'s and \emph{prior}'s means and log-variances.

\begin{table}[!t]
    \centering
    \resizebox{\columnwidth}{!}{
    \begin{tabular}{l|ccccc}
     \hline \toprule \noalign{\smallskip} 
         Hyperparameter & \textbf{ETH/UCY} & \textbf{SDD} & \textbf{STATS SportVU NBA} &  \textbf{TrajNet++} & \textbf{inD}\\ \midrule
         Optimizer & Adam  & Adam & Adam & SGD &  Adam \\
         Learning rate & $10^{-3}$ & $10^{-3}$ &  $10^{-3}$ & $3\times10^{-4}$ & $10^{-4}/5\times10^{-4}$\\
         Batch size & $16$ & $16$ & $32$ & $8$ & $16$\\
         Latent space size & $16$ & $16$ & $32$ & $16$ & $16$\\
         Warm-up epochs & $50$ & $50$ & - & $3$ & $50$ \\\bottomrule
    \end{tabular}}
    \caption{Main hyperparameters used to train both AC-VRNN and A-VRNN models on tested datasets.}
    \label{tab:params}
\end{table}

\begin{table}[!t]
\centering
\resizebox{\columnwidth}{!}{
\scalebox{0.8}{
\begin{tabular}{c|c}
\toprule
Module & Architecture\\ \toprule
Features extraction (trajectory) &  Linear (2, 64) $\rightarrow$ LeakyReLU $\rightarrow$ Linear(64, 64) $\rightarrow$ LeakyReLU                   \\ \midrule
Features extraction (belief map) & Linear (64, 64) $\rightarrow$ LeakyReLU                                                                      \\ \midrule
Prior                            & Linear(128, 64) $\rightarrow$ LeakyReLU                                                                    \\
Mean                             & Linear(64, 16)                                                                                              \\
Log-variance                     & Linear(61, 16)                                                                                              \\ \midrule
Encoder                          & Linear(192, 64) $\rightarrow$ LeakyReLU $\rightarrow$ Linear(64, 64) $\rightarrow$ LeakyReLU                  \\
Mean                             & Linear(64, 16)                                                                                              \\
Log-variance                     & Linear(64, 16)                                                                                               \\ \midrule
Latent space                           & Linear(16, 64) $\rightarrow$ LeakyReLU                                                                       \\ \midrule
Decoder                          & Linear(192, 64) $\rightarrow$ LeakyReLU $\rightarrow$ Linear(64, 64) $\rightarrow$ LeakyReLU               \\
Mean                             & Linear(64, 16) $\rightarrow$ HardTanH(-10, 10)                                                              \\
Log-variance                     & Linear(64, 16)                                                                                               \\ \midrule
Recurrence                       & GRU(128, 64, 1)                                                                                              \\ \midrule
Graph                            & GraphAttentionLayer(64, 64, hidden\_units=8, heads=4, $\alpha=0.2$) $\rightarrow$ BatchNorm1D $\rightarrow$ TanH \\ \toprule
\end{tabular}
}}
\caption{Detailed description of each module of our AC-VRNN architecture.}
\label{tab:backbone}
\end{table}

\section{Conclusion}
In this paper, we proposed a novel architecture for multi-future trajectory forecasting. Our framework uses VRNNs in a predictive setting. An attentive module includes interactions through a hidden state refinement process based on a graph neural network in an online fashion at a time step level. Finally, local belief maps encourage the model to follow a future displacement probability grid when the model is not confident about its prediction. We refer to our model as AC-VRNN and test it on several trajectory prediction datasets collected in different urban scenarios achieving the best performance compared to state-of-the-art methods.
Our future work will be towards a detailed analysis of long-term predictions in order to deal with more complex and uncertain scenarios. Furthermore, an interesting aspect would be to include into the model additional scene context (e.g., depth data or WiFi/BLT signals) in order to design a multi-modal architecture to gain the advantage of multiple modalities.
\section*{Acknowladgements}
Funded by the PRIN PREVUE - PRediction of activities and Events by "Vision in an Urban Environment" project (CUP E94I19000650001), PRIN National Research Program, MUR.

\bibliographystyle{model2-names}
\bibliography{references.bib}
\clearpage

\end{document}